\documentclass[11pt]{article}

\usepackage[final]{acl}
\usepackage{times}
\usepackage{latexsym}
\usepackage[T1]{fontenc}
\usepackage[utf8]{inputenc}
\usepackage{microtype}
\usepackage{inconsolata}
\usepackage{graphicx}
\usepackage{booktabs}
\usepackage{amsmath}
\usepackage{amssymb}
\usepackage{enumitem}
\usepackage{array}
\usepackage{xcolor}
\newcommand{\rev}[1]{#1}

\usepackage{pgfplots}
\pgfplotsset{compat=1.18}
\usepackage{tikz}
\usepackage{multirow}
\usepackage{hyperref}

\title{Policy Loopholes in Agent Evaluation:\\When Policy Ambiguity Masquerades as Agent Error}

\author{Hongliu Cao \\
  Amadeus SAS \\
  \small{
    \textbf{Correspondence:} \href{mailto:caohongliu@gmail.com}{caohongliu@gmail.com}
  }
}
\begin{document}
\maketitle

\begin{abstract}
Agent benchmarks evaluate policy compliance but assume each policy determines a unique correct action.
Natural-language policies can violate this assumption through silence, ambiguity, or contradiction, admitting multiple defensible readings that a single gold trajectory cannot capture. Auditing two $\tau^2$-bench domains, we develop a taxonomy of such policy loopholes and show that affected tasks produce unreliable scores: they lower scores across different models in different ways and make every model less consistent across repeated trials. A cross-domain comparison reveals that exploitability requires both policy ambiguity and tool permissiveness: when policy complexity exceeds what tools can enforce, agents resolve gaps inconsistently and scores become unreliable. Policy specification quality sets the ceiling on evaluation quality. Benchmark developers should audit policies before collecting gold annotations.
\end{abstract}

\section{Introduction}

Agent benchmarks, whether they score final outcomes~\citep{yao2024tau,barres2025tau2,zhou2023webarena,jimenez2024swebench} or audit procedural steps~\citep{levy2024stwebagentbench,li2025sopbench,ruan2024toolemu}, treat policy specifications as unambiguous ground truth.
They typically compare agent actions to a gold trajectory, add procedural checks over agent actions, or rely on tool guards to reject invalid operations.
We show that this assumption can break in practice.
Natural-language policies written for human employees are adequate for typical cases but can be ambiguous at decision boundaries, admitting multiple defensible readings that a single gold action cannot represent.
LLM agents lack the institutional context and supervisory access that human employees use to navigate such gaps. Hence, they resolve ambiguities differently: one model escalates, another proceeds conservatively, a third improvises a workaround.
Each response can be defensible under the policy text, while single-gold evaluation treats all but one as error, converting policy ambiguity into illusive capability gaps.

We call these ambiguous points \textbf{policy loopholes}: places where the policy text is silent, contradictory, or internally inconsistent, such that a competent agent could justify more than one action (formally defined in \S\ref{sec:related}).
For instance, one clause prohibits modifying basic economy flights while another permits cabin changes for all reservations. Agents disagree on whether upgrading cabin first makes the flight modifiable.
Existing benchmarks check agent compliance against the policy but rarely audit whether the policy itself determines a unique answer.
This leaves a blind spot: a policy-compliant agent can still disagree with the gold actions.

We propose a two-stage policy-specification audit that combines proactive clause analysis with reactive trace verification to systematically surface these gaps.
Our main contributions are:
\begin{enumerate}[leftmargin=*,nosep]
\item \textbf{A policy-specification audit methodology} for policy-grounded benchmarks, combining proactive clause analysis with reactive trace verification and a five-category loophole taxonomy: \rev{12 of 17 loopholes were surfaced proactively (Stage-1-only) and 5 were dually confirmed through both stages}. Although both domains contain policy gaps, only airline tasks show reduced scores while retail tasks are unaffected (\S\ref{sec:method}, \S\ref{sec:results}).
\item \rev{\textbf{A proposed two-factor mechanism}}: based on cross-domain analysis, we \rev{observe} that policy ambiguity becomes exploitable only when policy complexity exceeds what tool-level enforcement can close, \rev{explaining} the sharp domain split (\S\ref{sec:cross_domain}).
\item \textbf{A generalizability analysis} showing that loopholes are properties of policies, not individual tasks: $\tau^3$-bench's per-task fixes resolve symptoms but leave ambiguous clauses to generate new conflicts. Per-task patches proliferate and can contradict one another while policy-level fixes generalize across tasks. (\S\ref{sec:cross_domain}).
\end{enumerate}

\section{Related Work}
\label{sec:related}

We define a policy loophole as a point in a benchmark's policy document where the policy text is ambiguous, silent on an edge case, or admits multiple defensible readings, such that a competent agent could justify more than one course of action from the written text alone.
Loopholes are distinct from annotation conflicts, where a gold action is inadmissible under any reading of the policy.
In a loophole, the gold is one valid interpretation among several; in an annotation conflict, the gold contradicts the policy text.
Conflicts require re-annotation; loopholes require policy tightening or multi-gold evaluation (see Appendix~\ref{app:distinction} for a detailed comparison).
These loopholes are not artifacts of LLM misunderstanding; they are inherent to the policy specification itself.

Current agent benchmarks~\citep{yao2024tau,zhou2023webarena,liu2024agentbench,mialon2023gaia,jimenez2024swebench,xu2024theagentcompany} predominantly evaluate agents against unique gold outcomes.
A growing line of work adds procedural checks ~\citep{cao2026beyond}: ST-WebAgentBench~\citep{levy2024stwebagentbench}, SOPBench~\citep{li2025sopbench}, GuideBench~\citep{diao2025guidebench}, and INSURE-Dial~\citep{kulkarni-etal-2026-insure} score policy or guideline compliance; Beyond-IVR~\citep{balaji2026beyondivr} benchmarks business-adherence in customer support; ToolEmu~\citep{ruan2024toolemu} checks safe tool use; and AgentHarm~\citep{andriushchenko2024agentharm}, AgentDojo~\citep{debenedetti2024agentdojo}, and \citet{chang2025keep} test whether agents resist harmful or injected instructions.
Recent work also targets enforcement and failure detection: \citet{zwerdling2025towards} enforce policy adherence in agentic workflows, \citet{nakash2025effective} red-team policy-adherent agents, and \citet{rabinovich2025nearmiss} detect latent policy failures.
Despite recent advances, state-of-the-art methods still exhibit important limitations.
Benchmarks with the most realistic policies (prose with nested conditionals such as  $\tau^2$-bench) rely on the weakest verification: outcome-level matching against a single gold trajectory.
Benchmarks with stronger verification (executable guards, formal constraint solvers~\citep{winston2025solveraided,anand2026mantra}) require translating policies into formal specifications that may lose natural-language nuance, and process-level evaluation~\citep{gritta-etal-2026-process} still requires an unambiguous reference trajectory.
These limitations create a gap in which loopholes can emerge.
When policies are realistic enough to contain genuine ambiguity but verification checks only whether the outcome matches one annotator's reading, alternative interpretations register as failure regardless of their defensibility under the policy text.

Benchmark label errors are pervasive~\citep{northcutt2021pervasive}, and human disagreement often reflects genuine ambiguity rather than annotator error~\citep{pavlick2019inherent,plank2022problem,basile2021need}.
Recent work~\citep{bavaresco2025llms,bai2024mtbench101} examines how evaluation assumptions break down under scrutiny.
Ambig-SWE~\citep{vijayvargiya2026ambigswe} and CondAmbigQA~\citep{li2025condambigqa} treat underspecification as an explicit evaluation target, showing that apparent failures may stem from inherent ambiguity in user instructions or queries.
We extend this insight to policy specification: when the policy specification admits multiple defensible readings, both outcome-based and procedure-based evaluation inherit the ambiguity.
Just as statutory ambiguity in law produces circuit splits (different courts reaching different conclusions from identical text), policy ambiguity in benchmarks produces model splits: different models reaching different conclusions from identical policy, with only one arbitrarily deemed correct.

\section{Audit Framework}
\label{sec:method}

We propose a two-stage policy-specification audit for policy-governed benchmarks.

\paragraph{Stage 1: Proactive clause-level audit.}
Before examining any task outcomes, we analyze each policy clause and apply four diagnostic tests:
scope (does the clause cover this scenario?),
boundary (does it address edge cases?),
composition (does it compose unambiguously with other clauses?), and
assumption (does it rely on unstated presuppositions?).
Any clause failing one or more tests is flagged as a candidate loophole.

\paragraph{Stage 2: Reactive trace-level verification.}
For tasks where agents fail, we check whether (a)~the agent relied on a specific policy clause, (b)~that clause is genuinely ambiguous or silent, and (c)~a competent human agent could defensibly make the same decision.
Tasks meeting all three criteria are classified as loophole-affected (full criteria in Appendix~\ref{app:criteria}).

\paragraph{Loophole taxonomy.}
Applying these tests across both domains yields five categories of loopholes (full catalog in Appendix~\ref{app:catalog}).
A scope gap arises when the policy is entirely silent on a scenario.
A silent-on-edge loophole occurs when a clause covers the general topic but omits a specific boundary condition.
Conflicting clauses produce contradictory guidance when two individually clear rules apply to the same situation.
An implicit precondition arises when a rule depends on a concept the policy never defines.
Compositional ambiguity occurs when individual rules are each clear in isolation but the policy does not specify whether they can be applied in sequence.

\section{Experiments}
\label{sec:results}

We apply the audit framework to $\tau^2$-bench's airline (50 tasks) and retail (114 tasks) domains.
Trajectories come from GPT-5, Kimi-K2-Thinking, and Mistral-Large-3 (4 trials each), totaling 1,968 runs.
\rev{The three models represent three distinct providers and regions (OpenAI/US, Moonshot/China, Mistral/EU) and exhibit distinct disambiguation behaviors on loophole tasks, indicating that loophole sensitivity stems from benchmark policies rather than any individual model or training regime.}
\rev{All 17 loopholes were identified in Stage~1 from the policy text alone, before any agent trace was examined. Among which, 5 were additionally confirmed in Stage~2 on tasks with at least one failing trial. Those 5 loopholes account for seven affected tasks while the remaining 12 are attested by clause analysis only, suggesting that the two stages are complementary.} The full catalog can be found in Appendix~\ref{app:catalog}, where each receives an ID (L1--L16 for airline, R1--R2 for retail) referenced in the tables below.

\subsection{Airline: Exploitable Loopholes}

\begin{table}[t]
\centering
\small
\setlength{\tabcolsep}{3pt}
\begin{tabular}{@{}lp{2.4cm}ccc@{}}
\toprule
\textbf{Task} & \textbf{Loophole} & \rev{\textbf{GPT-5}} & \textbf{Kimi} & \rev{\textbf{Mistral}} \\
\midrule
2 & Sil.-on-Edge (L3/4) & 0/4 & 0/4 & 1/4 \\
5 & Sil.-on-Edge (L5) & 4/4 & 3/4 & 2/4 \\
13 & Scope Gap (L16) & 2/4 & 4/4 & 3/4 \\
27 & Sil.-on-Edge (L3/4) & 0/4 & 0/4 & 0/4 \\
29 & Scope Gap (L16) & 0/4 & 2/4 & 3/4 \\
32 & Comp.\ Ambig.\ (L6) & 1/4 & 0/4 & 0/4 \\
45 & Comp.\ Ambig.\ (L6) & 2/4 & 3/4 & 1/4 \\
\midrule
\multicolumn{2}{@{}l}{\textbf{Loophole avg.}} & 32.1\% & 42.9\% & 35.7\% \\
\multicolumn{2}{@{}l}{\textbf{Non-loophole avg.}} & 64.5\% & 48.3\% & 40.7\% \\
\bottomrule
\end{tabular}
\caption{Per-task success on seven loophole-affected airline tasks (successes / 4 trials). Bottom rows compare average success on loophole vs.\ non-loophole tasks.}
\label{tab:cross_model}
\end{table}
Seven of the 50 airline tasks are affected by at least one loophole.
\rev{The clearest signal is inconsistency: the same type of policy gap yields drastically different outcomes across tasks: Task~27 receives 0/12 successful runs, whereas Task~5 receives 9/12. The difference is not the nature of the gap itself, but which interpretation the annotator selected as gold actions.
Within-task cross-trial variance on loophole tasks is $1.62\times$ higher than on non-loophole tasks (0.110 vs.\ 0.068 across 21 and 129 model-task cells respectively), quantifying the evaluation-reliability cost \rev{(details in Appendix~\ref{app:stats})}}.

Loophole tasks achieve an average success rate of 37\%, compared with 51\% for non-loophole tasks (Table~\ref{tab:cross_model}). However, this aggregate masks heterogeneous per-model gaps: 
GPT-5 shows a 32.4 percentage-point loophole penalty,
In contrast, Kimi and Mistral show smaller performance differences (around 5pp) between loophole and non-loophole tasks. Notably, GPT-5 is also the strongest performer on non-loophole tasks. If loophole tasks were simply more difficult, stronger models would be expected to suffer less, not more. Instead, the observed pattern suggests that policy ambiguity creates evaluation artifacts that disproportionately penalize certain disambiguation strategies.

\rev{Counterfactual re-aggregation quantifies the benchmark-wide impact: removing the 7 loophole tasks lifts GPT-5's overall success rate by 4.5\,pp, from 60.0\% to its non-loophole rate of 64.5\%, versus only 0.8 and 0.7\,pp for Kimi and Mistral, and widens GPT-5's lead over Kimi from +12.5 to +16.2\,pp. The current benchmark therefore suppresses the strongest model's score.
However, since Stage~1 flagged 15 valid airline loopholes but only 7 are exercised by current gold trajectories, the 7-task count is a lower bound on the policy problem. Greater task diversity within the same policy space may reveal additional loopholes.}

Three patterns of loophole failure recur.
First, silent-on-edge cases: in Tasks~2 and 27, the compensation clause says ``after confirming facts and changing or cancelling''; agents consistently read ``changing or cancelling'' as a prerequisite and withhold the certificate, but the gold encodes the lenient reading (1/12 and 0/12 success respectively).
Second, scope gaps: in Tasks~13 and 29, the policy prohibits modifying basic economy flights or changing destinations, but is silent on whether cancel-and-rebook achieves the same effect through individually permitted actions. Some agents find this workaround while others correctly refuse, and both behaviors are defensible.
Third, compositional ambiguity: in Tasks~32 and 45, the policy allows basic economy to upgrade cabin and allows non-basic-economy to modify flights, but is silent on whether these compose sequentially. Agents that attempt the two-step workaround (upgrade, then modify) split with those that refuse, producing inconsistent scores.

The core problem is not low or high success but unreliable evaluation.
\rev{Tasks with high success are equally misleading: Task~5 (75\% overall) still produces cross-model divergence (GPT-5 4/4, Kimi 3/4, Mistral 2/4), and it was flagged proactively from the policy text before any traces were examined.
Together with the capability-inverted gap (the strongest model suffers the largest penalty) and the zero-success signature (Task~27, 0/12), these three lines of evidence are inconsistent with a task-difficulty confound and consistent with policy ambiguity driving the score depression.
Loophole sensitivity tracks whether a model's default disambiguation strategy happens to match the annotator's unstated preference, not model quality.}
Models resolve ambiguity differently: \rev{Kimi-K2-Thinking is the most conservative, refusing workarounds that other models attempt. GPT-5 most often explores workarounds} and Mistral-Large-3 is the least consistent. No single strategy dominates (Appendix~\ref{app:patterns}).
Full case studies can be found in Appendix~\ref{app:cases}.

\subsection{Retail and Cross-Domain Analysis}
\label{sec:cross_domain}

Applying the same two-stage audit to the retail domain (114 tasks, 1,368 runs), Stage~1 identified two policy ambiguities (R1--R2) and Stage~2 confirmed zero exploitable tasks.
R1 concerns a minor ambiguity in the modify-once constraint: the policy states that tools ``can only be called once per order'' but does not specify whether ``once'' means one successful execution or one attempt, including failed calls with incorrect parameters.
R2 concerns the lifecycle of a modified order: the policy is silent on whether an order whose status changed to \texttt{"pending (items modified)"} can later be cancelled or returned.
Both are genuine textual gaps, but in both cases the tools close them: modifying an order moves its status out of \texttt{"pending"}, and all subsequent write calls are rejected because the precondition no longer holds.
The agent never gets to exercise either ambiguous interpretation.

\paragraph{Why airline but not retail?}
The two domains differ in both policy complexity and tool-level enforcement, and these factors appear to reinforce one another. However, because the airline domain is simultaneously more complex and less constrained by tool validation, their individual effects cannot be disentangled from a comparison of only two domains. Our mechanism-level hypothesis is that complex policies are harder to encode as tool constraints, making enforcement more dependent on agent reasoning. While this explanation is consistent with the observed pattern, confirming it would require either additional domains or controlled manipulations of policy complexity and tool enforcement.

Airline policies contain interconnected and highly compositional eligibility rules. For example, cancellation eligibility is defined by four disjunctive conditions, while compensation requires both a qualifying complaint and a change or cancellation event, together with membership, insurance, or business-class status. In addition, multiple workflows share state through variables such as cabin class, insurance coverage, and membership status. This complexity makes exhaustive enforcement through tool-level constraints impractical. The policy explicitly warns that ``the API does not check'' cancellation or compensation eligibility, leaving the agent as the sole enforcer.
Because reservations retain their status after writes, multi-step compositions (upgrade then modify, cancel then rebook) are possible, and different agents enforce the ambiguous rules differently.

Retail eligibility is binary: each workflow is gated by a single status field (\texttt{"pending"} or \texttt{"delivered"}), with no cross-workflow dependencies.
This simplicity makes tool enforcement feasible: retail tools embed 32 policy-level validation checks and move orders into terminal statuses that block further writes, creating a one-shot design where multi-step sequences cannot arise.
The policy gaps still exist in the text, but the tools close them before any agent behavior can diverge.
Table~\ref{tab:two_layer} summarizes these structural differences.
In summary, when policy complexity exceeds what tools can enforce, loopholes become exploitable and evaluation becomes unreliable.

\begin{table}[t]
\centering
\small
\begin{tabular}{@{}p{4.7cm}cc@{}}
\toprule
\textbf{Factor} & \textbf{Airline} & \textbf{Retail} \\
\midrule
\multicolumn{3}{@{}l}{\textit{Policy complexity}} \\
\quad Cataloged loopholes & \rev{15} & 2 \\
\quad Conditional clauses (if) & 24 & 13 \\
\quad Disjunctive eligibility rules & 2 & 0 \\
\quad Cross-workflow dependencies & Dense & Minimal \\
\midrule
\multicolumn{3}{@{}l}{\textit{Tool enforcement}} \\
\quad ``API does not check'' disclaimers & 2 & 0 \\
\quad Tool-level input validation checks & 18 & 32 \\
\quad Terminal statuses blocking further action & 1 & 4 \\
\midrule
\multicolumn{3}{@{}l}{\textit{Outcome}} \\
\quad Triggered loophole tasks & 7 & 0 \\
\bottomrule
\end{tabular}
\caption{Policy complexity and tool enforcement jointly determine loophole exploitability. Simple policies are encodable in tools; complex policies are not, leaving the agent as the sole enforcer.}
\label{tab:two_layer}
\end{table}

\paragraph{\rev{Supporting evidence from $\tau^3$-bench revisions.}}
Concurrent with our work, the $\tau^3$-bench release \rev{independently identified and revised all seven loophole-affected tasks within a set of 29/50 airline tasks it changed (counted from pinned commits; see Appendix~\ref{app:tau3}).
Under a null of independent selection, the probability that all~7 fall inside that set by chance is 1.6\% (hypergeometric, $N{=}50$, $K{=}29$, $n{=}7$).
The overlap is not only statistical but terminological: $\tau^3$'s explicit ``Policy Loophole Prevention'' category contains four tasks (13, 29, 32, 45), all four in our seven.
The remaining three (Tasks~2, 5, 27) $\tau^3$ classifies under ``Incorrect Action'' or ``Ambiguous User Instructions,'' which are different top-level labels for the same underlying ``not uniquely correct'' judgment that our loophole/conflict taxonomy unifies.}
However, every fix modifies user instructions or gold actions; the policies themselves remain unchanged, and the underlying ambiguities persist.
The inconsistency is sharpest for L6: $\tau^3$ hard-coded opposite answers for two tasks sharing the same clause (Task~32: workaround permitted; Task~45: workaround refused), each resolved via per-task user instructions rather than a single clarifying rule.
These fixes address symptoms, not the disease: they protect seven specific tasks but leave the ambiguous clauses intact for future tasks.
Fixing ambiguities through user instructions rather than policy clarifications forces every future annotator to reverse-engineer the maintainers' intent.
A single clarifying policy sentence would generalize without contradictory annotations (see Appendix~\ref{app:tau3}).

\section{Conclusion}

Policy specification quality is the ceiling on evaluation quality.
In this paper, we introduce a two-stage audit combining proactive clause analysis with reactive trace verification to surface policy gaps.
Auditing two $\tau^2$-bench domains, we find that both contain ambiguities admitting multiple defensible readings, but only airline tasks show reduced scores. Based on the cross-domain analysis, we identify that when policy complexity exceeds what tools can enforce, agents resolve ambiguities inconsistently.
These findings show that policy ambiguity is not merely an annotation artifact; it is a fundamental threat to the validity of procedural agent evaluation. Evidence from $\tau^3$-bench further reinforces this conclusion. Per-task patches leave the underlying ambiguity unchanged and can even create contradictory gold answers, whereas policy-level corrections resolve the ambiguity once and generalize across all affected tasks.
Benchmark developers should audit policies before collecting gold annotations, report loophole-affected and loophole-free scores separately, and accept multiple valid outcomes where the policy is genuinely underspecified.
Without these steps, benchmark scores risk measuring policy underspecification rather than agent capability.

\section*{Limitations}

Our audit covers two domains from one benchmark; the cross-domain contrast strengthens generalizability, but \rev{additional domains are needed to confirm the proposed two-factor mechanism, since complexity and enforcement covary in our data}.
The proactive audit is AI-assisted but human-driven; fully automated clause-level analysis could help in larger policy documents.
Scaling the audit further will require automating both loophole detection and validation; multi-agent LLM judge frameworks~\citep{cao2025multi} offer a potential path for the latter.
The reasonableness test involves subjective judgment, which we mitigate with explicit criteria (Appendix~\ref{app:criteria}).
Loophole severity depends on both policy structure and tool design; future work should quantify this interaction across more domains and benchmark families.
Our taxonomy targets prose-style policies with nested conditionals; benchmarks using graph-structured SOPs or formal constraint specifications may exhibit different loophole distributions.

\section*{Ethical Considerations}

This work audits a publicly released benchmark with synthetic users and simulated environments; it does not involve human subjects, private data, or deployment of AI systems.
While our taxonomy could in principle help identify exploitable ambiguities in real-world policies, \rev{the specific loopholes we enumerate are drawn from a synthetic benchmark}, and the primary mitigation we recommend (tightening policy specifications) directly reduces any such attack surface.
Our audit is intended to improve evaluation quality, not to enable exploitation of policy gaps in deployed systems.

\paragraph{Use of AI Assistance.} Generative AI tools were used to assist with policy clause analysis during the audit (identifying candidate ambiguities for human review), as well as grammar correction and proofreading. All loophole classifications, taxonomy decisions, and analytical claims were from the authors, who take full responsibility for the content and integrity of the submitted work.

\bibliography{custom}

\appendix

\section{Audit Methodology and Classification Criteria}
\label{app:method}
\label{app:criteria}

\subsection{Task-Document Generation}
For each task, we compile a structured document containing: task description, user instruction, database state (reservation records, user profiles, order details), the complete gold action sequence, and the policy text.
This document is self-contained, enabling policy-check analysis without access to agent trajectories.

\subsection{Proactive Audit Procedure}
We analyze each policy document clause by clause and identify ambiguity via four tests:
\begin{itemize}[nosep,leftmargin=*]
\item \textbf{Scope}: Does the clause cover this scenario, or is it silent?
\item \textbf{Boundary}: Does the clause address edge cases (e.g., zero-cost modifications, user declining an offer)?
\item \textbf{Composition}: When multiple clauses apply, do they compose unambiguously?
\item \textbf{Assumption}: Does the clause rely on unstated common-sense assumptions?
\end{itemize}

\subsection{Per-Task Loophole Classification}
For each task flagged during the proactive audit or \rev{with at least one failing trial across model-trial combinations}, we apply the three-criteria test described in \S\ref{sec:method}.
We also apply exclusion criteria: if the gold action is inadmissible under any reading of the policy, the task is classified as an annotation conflict rather than a loophole.

\subsection{\rev{Statistical Procedures}}
\label{app:stats}

\rev{\paragraph{Per-model gap confidence intervals.}
Each model contributes 28 loophole-task trials (7 tasks $\times$ 4) and 172 non-loophole trials (43 tasks $\times$ 4). We resample the two strata independently at the trial level, with replacement and at their original sizes, recompute the gap between the two mean success rates, and repeat 10{,}000 times. Reported intervals are the 2.5th and 97.5th percentiles of the resulting distribution (percentile method, seed 42).}

\rev{\paragraph{Within-task cross-trial variance.}
For each model-task cell we compute the population variance ($\mathrm{ddof}=0$) of the four binary trial outcomes, then average across cells within each group. This gives 21 loophole cells (7 tasks $\times$ 3 models) and 129 non-loophole cells (43 $\times$ 3). The metric captures how far a cell sits from a deterministic outcome: it is 0 when all four trials agree and reaches its maximum at 0.25 when they split evenly, so a higher average means the same model reaches different conclusions on the same task across trials.}

\subsection{Per-Loophole Decision Rules}

We apply the following decision rules to classify tasks as loophole-affected.
These are designed to be reproducible by independent auditors.

\paragraph{Compensation prerequisite (L3/L4).}
The clause ``after confirming facts and changing or cancelling'' is classified as ambiguous because ``after'' can denote temporal sequencing (prerequisite) or typical co-occurrence (independent). If the gold issues a certificate without a completed change/cancel, the task is loophole-affected.

\paragraph{Disjunctive eligibility (L5).}
If the compensation eligibility has OR conditions and the gold anticipates only one path, any task where an agent finds a valid alternative path is loophole-affected.

\paragraph{User autonomy (L13).}
If the gold assumes user acceptance of a cost/action and the policy does not specify fallback behavior on user decline, the task is loophole-affected.

\paragraph{Temporal ambiguity (L12).}
If a reservation's flight dates are in the past but its status field does not explicitly mark it as ``flown'' or ``completed,'' the cancellability is ambiguous and the task is loophole-affected.

\paragraph{Compositional ambiguity (L6).}
If the gold requires a specific ordering of API calls (sequential) but the policy does not mandate ordering and a single-call approach produces an equivalent final state, the task is loophole-affected.

\paragraph{Cancel-and-rebook bypass (L16).}
If the policy prohibits modification (basic economy) or destination change, but is silent on whether cancel+rebook achieves the same effect through individually permitted actions, the task is loophole-affected.

\paragraph{Exclusion criteria.}
A task is not a loophole if: (a)~the gold action is inadmissible under any policy reading (annotation conflict); (b)~the failure is caused by simulator deviation rather than policy ambiguity; or (c)~the agent makes a clear procedural error unrelated to the ambiguity.

\section{Loophole vs.\ Annotation Conflict}
\label{app:distinction}

Table~\ref{tab:distinction} summarizes the key distinctions.

\begin{table}[!ht]
\centering\small
\setlength{\tabcolsep}{3pt}
\begin{tabular}{@{}p{1.9cm}p{2.4cm}p{2.4cm}@{}}
\toprule
\textbf{Dimension} & \textbf{Loophole} & \textbf{Conflict} \\
\midrule
Gold action & One valid reading among several & Inadmissible under any reading \\
Policy text & Ambiguous or silent & Clear but contradicted by gold \\
Fix required & Policy tightening or multi-gold & Re-annotation \\
Agent blamed? & Unfairly; defensible path & Unfairly; correct refusal \\
\bottomrule
\end{tabular}
\caption{Loopholes vs.\ annotation conflicts: key distinctions.}
\label{tab:distinction}
\end{table}

\section{Full Loophole Catalog}
\label{app:catalog}

Tables~\ref{tab:airline_catalog} and~\ref{tab:retail_catalog} list every ambiguity surfaced by the audit (airline L1--L16, retail R1--R2). \rev{Of these 18 entries, 17 meet our loophole definition; L15 is an annotation conflict and is excluded from the loophole count.}

\rev{Two columns need definition. Observed Effect records whether an ambiguity produced evaluation mismatches in our experiments: \checkmark\ a direct effect on scores, Indirect a contributing but not sole cause, Borderline marginal evidence, Untested no current task exercising the gap, and Conflict an annotation conflict rather than a loophole. Disc.\ (Discovery) marks whether an entry was identified in Stage~1 only (S1) or confirmed through both stages (S1+S2).} Key policy excerpts follow.

\begin{table*}[tp]
\centering
\small
\begin{tabular}{@{}clp{5.0cm}llc@{}}
\toprule
\textbf{\#} & \textbf{Type} & \textbf{Ambiguity} & \rev{\textbf{Obs.\ Effect}} & \rev{\textbf{Disc.}} & \textbf{Tasks} \\
\midrule
L1 & Scope Gap & ``Business flight'': booked as business, or currently business after upgrade? & \rev{Untested$^\dagger$} & \rev{S1} & \rev{---} \\
L2 & Scope Gap & ``Other reasons'' for cancellation: what qualifies? & Untested & \rev{S1} & --- \\
L3 & Silent-on-Edge & ``Wants to change or cancel'': prerequisite or typical scenario? & \checkmark & \rev{S1+S2} & 2, 27 \\
L4 & Silent-on-Edge & What if user wants compensation but NOT to change/cancel? & \checkmark & \rev{S1+S2} & 2, 27 \\
L5 & Silent-on-Edge & Disjunctive eligibility: policy silent on whether agent must proactively check all OR paths when trigger clause is borderline & \checkmark & \rev{S1+S2} & 5 \\
L6 & Comp.\ Ambig. & After cabin upgrade, is former basic-economy fully modifiable? & \checkmark & \rev{S1+S2} & 32, 45 \\
L7 & Implicit Prec. & ``Kept'' segments: explicitly unchanged or auto-kept? & Untested & \rev{S1} & --- \\
L8 & Silent-on-Edge & Zero price difference: is payment method still required? & Untested & \rev{S1} & --- \\
L9 & Implicit Prec. & ``Within last 24 hrs'': $\leq$24h or $<$24h? From when? & Borderline & \rev{S1} & --- \\
L10 & Comp.\ Ambig. & After cabin upgrade, do baggage allowances auto-update? & Untested & \rev{S1} & --- \\
L11 & Implicit Prec. & ``Transfer if cannot be handled'': impossible vs.\ policy violation? & Untested & \rev{S1} & --- \\
L12 & Scope Gap & Delayed/on-time flight ``cannot be booked'': but cancellable? & Untested & \rev{S1} & --- \\
L13 & Implicit Prec. & ``The user is required to pay for the difference''\footnotemark; silent on user decline & Untested & \rev{S1} & --- \\
L14 & Silent-on-Edge & Compensation ``per passenger'': per cancelled flight or reservation? & Untested & \rev{S1} & --- \\
L15 & Conflicting & Insurance ``health or weather'': applies to eligibility or refund? & Conflict$^*$ & \rev{---} & 44 \\
L16 & Scope Gap & Cancel+rebook as bypass for modification restrictions (basic economy, destination) & \checkmark & \rev{S1+S2} & 13, 29 \\
\bottomrule
\end{tabular}
\caption{Complete airline policy loophole catalog (L1--L16); \rev{column values are defined in the text. Of the 16 entries, 15 are loopholes.} $^*$Task~44 triggers L15 but its gold is inadmissible under any reading, so L15 is an annotation conflict \rev{carrying no discovery label and excluded from the 17-loophole count. $^\dagger$Task~7 turns on L1's clause, but its gold both modifies a basic-economy reservation and cancels reservations meeting none of the four cancellation conditions; we therefore treat L1 as untested and address Task~7 in companion work.}}
\label{tab:airline_catalog}
\end{table*}
\footnotetext{The policy requires payment but does not specify what happens if the user declines after being informed of the price.}

\begin{table*}[tp]
\centering
\small
\begin{tabular}{@{}clp{5.0cm}llc@{}}
\toprule
\textbf{\#} & \textbf{Type} & \textbf{Ambiguity} & \rev{\textbf{Obs.\ Effect}} & \rev{\textbf{Disc.}} & \textbf{Tasks} \\
\midrule
R1 & Silent-on-Edge & ``Exchange or modify order tools can only be called once per order'': does ``called once'' mean one successful execution or one attempt (including failed calls with wrong parameters)? Tool enforces one attempt, but policy text is ambiguous & Untested & \rev{S1} & --- \\
R2 & Comp.\ Ambig. & After modify, status becomes ``pending (item modified)''; policy is silent on whether it can be returned once delivered & Untested & \rev{S1} & --- \\
\bottomrule
\end{tabular}
\caption{Retail policy loophole catalog (R1--R2). Other candidates (partial cancellation, identical-item exchange, out-of-stock fallback) were clear in the policy text or not genuine ambiguities.}
\label{tab:retail_catalog}
\end{table*}

\paragraph{Key Policy Excerpts.}
\label{app:policy}
Below are verbatim clauses underlying the ambiguities cataloged above.

\paragraph{Compensation clause (airline).}
\begin{quote}
\small\itshape
``If the user complains about delayed flights in a reservation and wants to change or cancel the reservation, the agent can offer a certificate as a gesture after confirming the facts and changing or cancelling the reservation, with the amount being \$50 times the number of passengers.''
\end{quote}

\paragraph{Compensation eligibility (airline).}
\begin{quote}
\small\itshape
``Only compensate if the user is a silver/gold member or has travel insurance or flies business.''
\end{quote}

\paragraph{Basic economy modification (airline).}
\begin{quote}
\small\itshape
``Basic economy flights cannot be modified.'' + ``All reservations, including basic economy, can change cabin without changing the flights.''
\end{quote}

\paragraph{Cancellation eligibility (airline).}
\begin{quote}
\small\itshape
``Flight can be cancelled if any of the following is true: The booking was made within the last 24 hrs; The flight is cancelled by airline; It is a business flight; The user has travel insurance and the reason for cancellation is covered by insurance.''
\end{quote}

\paragraph{Modify-once constraint (retail).}
\begin{quote}
\small\itshape
``Exchange or modify order tools can only be called once per order. Be sure that all items to be changed are collected into a list before making the tool call!!!''
\end{quote}

\section{Case Studies}
\label{app:cases}
\label{sec:cases}

Each case study includes the relevant database state, gold actions, the violated policy clause, what a policy-compliant agent should do under each reading, and cross-model outcomes.

\subsection{Case Study 1: Cancel-and-Rebook Bypass (Tasks~13 \& 29, Airline)}

\paragraph{Database state.}
Task~13: \rev{User wants to change the destination of a basic economy return leg from ATL, from LAX to LAS,} willing to pay up to \$100 and upgrade to economy if needed. \rev{Two earlier segments of the same reservation have already been flown.}
Task~29: \rev{User wants to replace a DTW--LGA round trip with nonstop DTW--JFK flights on the same dates, plus one checked bag.}

\paragraph{Gold actions.}
Task~13: \texttt{transfer\_to\_human\_agents} (agent should refuse the modification).
Task~29: \texttt{update\_reservation\_flights} + \texttt{update\_reservation\_baggages}.

\paragraph{Ambiguous clause (L16).}
``Basic economy flights cannot be modified.'' + \rev{``Other reservations can be modified without changing the origin, destination, and trip type.''}
The policy is silent on whether cancelling and rebooking a new reservation achieves the same effect as the prohibited modification.

\paragraph{Two defensible readings.}
\begin{enumerate}[nosep,leftmargin=*]
\item \textbf{Spirit-of-the-rule}: Cancel+rebook is a modification workaround and should be refused.
\item \textbf{Letter-of-the-rule}: Cancel and rebook are separate permitted actions; the policy only restricts the \texttt{update\_reservation} tool.
\end{enumerate}
\rev{Whether the second reading is available at all is itself unsettled, for two separate reasons. First, the policy lists four conditions under ``flight can be cancelled if any of the following is true,'' yet one condition is travel insurance, which it elsewhere defines as enabling a refund. The list can therefore be read as gating cancellation, or as gating only the refund that follows it. GPT-5 takes both positions within Task~29, concluding in one trial that ``I'm not able to cancel your current reservation under policy'' and stating in another that ``cancellation for a refund is allowed only if'' those conditions hold. Second, in Task~13 the clause ``if any portion of the flight has already been flown'' never says whether ``the flight'' is the reservation or the segment being changed. Mistral reads it both ways, in one trial declining to ``modify or cancel the remaining flights'' and in another offering to cancel only the return portion.}

\paragraph{Agent behavior.}
\rev{All three models raise cancel-and-rebook with the user on Task~13, and in one GPT-5 trial the user explicitly selects it before the agent ultimately transfers.}
Kimi-K2-Thinking transfers in all 4 trials.
\rev{GPT-5 transfers in 2 and books a separate reservation in the other 2, scoring 0 both times. Mistral matches the gold in 3 of 4 trials.}
\rev{On Task~29 GPT-5 declines to modify and books a parallel reservation in 3 of 4 trials. It scores 0 in all four, although the evaluator awards full credit in every trial for what it tells the user. Kimi and Mistral modify the reservation directly and are scored correct in 2 and 3 trials respectively.}
\rev{The gold expectations are thus opposite across the two tasks: Task~13 rewards refusal, Task~29 rewards the modification} (Table~\ref{tab:case1}).

\begin{table}[!ht]
\centering\small
\begin{tabular}{@{}lcc@{}}
\toprule
\textbf{Model} & \textbf{Task 13} & \textbf{Task 29} \\
\midrule
GPT-5 & 2/4 & 0/4 \\
Kimi-K2-Thinking & 4/4 & 2/4 \\
Mistral & 3/4 & 3/4 \\
\bottomrule
\end{tabular}
\caption{Case~1 (Tasks~13 \& 29): same scope gap, opposite gold expectations.}
\label{tab:case1}
\end{table}

\subsection{Case Study 2: Compensation Prerequisite (Task~2, Airline)}

\paragraph{Database state.}
\rev{User Noah Muller complains about a delayed flight in his most recent reservation. He holds two reservations, so the gold has the agent inspect both before compensating on the affected one.}

\begin{small}
\begin{verbatim}
User noah_muller_9847:
  membership: gold
  reservations: [SDZQKO, 4OG6T3]

Reservation 4OG6T3 (BOS -> LAS):
  cabin: basic_economy
  passengers: 1
  insurance: yes
  flights: HAT006 (landed),
           HAT018 (delayed),
           HAT040, HAT253
\end{verbatim}
\end{small}

\rev{Compensation eligibility is satisfied twice over, since the user is a gold member and the reservation carries insurance, so the only question is whether the trigger clause permits a certificate at all.}

\paragraph{Gold actions.}
Gold expects \texttt{send\_certificate(amount=50)}.

\paragraph{Ambiguous clause.}
``If the user complains about delayed flights \rev{in a reservation} and wants to change or cancel the reservation, the agent can offer a certificate \rev{as a gesture} after confirming the facts and changing or cancelling the reservation, with the amount being \$50 \rev{times} the number of passengers.''

\paragraph{Two defensible readings.}
\begin{enumerate}[nosep,leftmargin=*]
\item \textbf{Strict}: ``changing or cancelling'' is a prerequisite; certificate requires completed change/cancel first.
\item \textbf{Lenient} (gold): certificate is independent of change/cancel; ``after'' describes typical but not required sequencing.
\end{enumerate}

\paragraph{Agent behavior.}
\rev{GPT-5 reads the clause strictly in all 4 trials and states the prerequisite explicitly, in one trial telling the user ``I can only issue a travel certificate for a delayed flight if the reservation is changed or cancelled due to the delay'' and in another that ``compensation for delays is only available when we change or cancel the affected reservation at your request.'' Mistral instead reads the clause leniently in 2 trials, issuing a certificate while the reservation stands: once for \$50, matching the gold, and once for \$200, having applied the \$100 cancelled-flight rate to a delay and counted the passengers of the wrong reservation. Kimi-K2-Thinking is uninformative here, as none of its 4 runs reached the compensation decision at all: three terminated on repeated tool errors and one on missing data, and none mentions compensation} (Table~\ref{tab:case2}).

\begin{table}[!ht]
\centering\small
\begin{tabular}{@{}lcccc c@{}}
\toprule
\textbf{Model} & T1 & T2 & T3 & T4 & \textbf{Sum} \\
\midrule
GPT-5 & 0 & 0 & 0 & 0 & 0/4 \\
Kimi-K2-Thinking & 0 & 0 & 0 & 0 & 0/4 \\
Mistral & 0 & \rev{1} & 0 & \rev{0} & 1/4 \\
\bottomrule
\end{tabular}
\caption{Cross-model outcomes for Case~2 (Task~2). \rev{Ten of the 12 runs issue no certificate, consistent with the strict reading; Mistral issues one twice, matching the gold's lenient reading once. Kimi's four runs ended on repeated tool errors.}}
\label{tab:case2}
\end{table}

\subsection{Case Study 3: Compensation Prerequisite Variant (Task~27, Airline)}

\paragraph{Database state.}
User Ethan Martin complains about delayed flight HAT039. Same compensation clause applies.

\paragraph{Gold actions.}
Gold expects \texttt{send\_certificate(amount=150)} (3 passengers $\times$ \$50).

\paragraph{Ambiguity.}
Same prerequisite ambiguity as Task~2 (L3/L4). The agent also must call \texttt{get\_user\_details} to check membership, but if the agent's reading excludes compensation (because user declined change/cancel), checking membership is unnecessary. No run scores above zero (Table~\ref{tab:case3}).

\rev{\paragraph{Agent behavior.}
One GPT-5 trial shows the cost of the strict reading most directly. Having concluded that a change is a prerequisite, the agent changes the flights and then issues the certificate for \$150, the exact gold amount. The evaluator records both gold actions as matched, \texttt{get\_user\_details} and \texttt{send\_certificate(amount=150)}, yet the run scores zero because the flight change it made to satisfy the prerequisite leaves the database diverging from a gold that assumes no change. An agent can therefore produce every action the gold requires and still fail, because the two readings disagree about what else must happen. The remaining GPT-5 trials withhold the certificate entirely. Mistral issues \$50 in one trial, reading the clause leniently but pricing it for one passenger rather than three. Kimi's first three runs terminated on repeated tool errors.}

\begin{table}[!ht]
\centering\small
\begin{tabular}{@{}lcccc c@{}}
\toprule
\textbf{Model} & T1 & T2 & T3 & T4 & \textbf{Sum} \\
\midrule
GPT-5 & 0 & 0 & 0 & 0 & 0/4 \\
Kimi-K2-Thinking & 0 & 0 & 0 & 0 & 0/4 \\
Mistral & 0 & 0 & 0 & 0 & 0/4 \\
\bottomrule
\end{tabular}
\caption{Cross-model outcomes for Case~3 (Task~27). \rev{No run scores above zero, including one GPT-5 trial in which both gold actions matched but an accompanying flight change, required under the strict reading, broke the database comparison.}}
\label{tab:case3}
\end{table}

\subsection{Case Study 4: Disjunctive Eligibility (Task~5, Airline)}

\paragraph{Database state.}
User Mei Brown has a business-class reservation with a delayed flight. Her membership is Regular (not Silver/Gold).

\paragraph{Gold actions.}
Gold expects only \texttt{get\_user\_details} and asserts ``Agent does not offer any compensation.'' The gold checks membership (Regular, not Gold) and concludes no compensation is warranted.

\paragraph{Ambiguity (L5).}
Compensation requires that the user ``wants to change or cancel'' (trigger clause) AND meets eligibility (``silver/gold member OR travel insurance OR flies business'').
The user flies business (eligible via the third disjunct) but does not explicitly request a change or cancel, so the trigger clause is borderline.
The gold checks only membership, finds Regular, and concludes no compensation, reaching the right outcome via an incomplete reasoning path.
The policy is silent on whether an agent should proactively offer compensation to an eligible user when the trigger clause is not clearly satisfied.

\paragraph{Agent behavior.}
GPT-5 matches the gold in all 4 trials (does not offer compensation). \rev{Kimi-K2-Thinking matches 3/4 but once issues a \$200 certificate, correctly priced at \$50 for the four passengers, after first changing the flights. Mistral offers compensation in 2/4 trials, in both cases having accepted the user's false claim of Gold status.} Agents that proactively check all eligibility disjuncts and offer compensation follow a defensible reading of the policy but are scored as failures (Table~\ref{tab:case4}).

\begin{table}[!ht]
\centering\small
\begin{tabular}{@{}lcccc c@{}}
\toprule
\textbf{Model} & T1 & T2 & T3 & T4 & \textbf{Sum} \\
\midrule
GPT-5 & 1 & 1 & 1 & 1 & 4/4 \\
Kimi-K2-Thinking & 1 & 1 & \rev{0} & \rev{1} & 3/4 \\
Mistral & 1 & 0 & \rev{0} & \rev{1} & 2/4 \\
\bottomrule
\end{tabular}
\caption{Cross-model outcomes for Case~4 (Task~5). High success means matching the gold's narrow reading (no compensation); agents that offer compensation are penalized, although the user flies business and so qualifies under the eligibility clause.}
\label{tab:case4}
\end{table}

\subsection{Case Study 5: Cabin Upgrade Unlocking Modification (Task~32, Airline)}

\paragraph{Database state.}
\rev{User Ivan Rossi wants to change an upcoming basic-economy itinerary (reservation OWZ4XL, EWR$\to$MIA$\to$LAX, 3 passengers) to a nonstop on the same day. He does not ask for a cabin upgrade; he states only that he is willing to upgrade to economy if the agent tells him the ticket is basic economy.}

\paragraph{Gold actions.}
\rev{Gold expects two separate \texttt{update\_reservation\_flights} calls. The first passes \texttt{cabin=economy} while re-submitting the original flights (HAT202, HAT232), effecting the upgrade without changing the itinerary; the second switches to the nonstop (HAT041). The gold thus treats the upgrade-then-modify sequence as valid.}

\paragraph{Ambiguity (L6).}
The policy says ``basic economy flights cannot be modified'' but also that ``all reservations, including basic economy, can change cabin \rev{without changing the flights}.''
\rev{The trailing qualifier is itself scope-ambiguous. Under a constraint reading, cabin may be changed only so long as the flights are not, which forecloses the workaround. Under a descriptive reading, it merely observes that a cabin change need not involve a flight change, leaving the sequence permitted. The gold's own first call satisfies even the constraint reading, changing cabin while leaving the flights untouched, so the two-step path is assembled entirely from individually compliant calls.}
If the user upgrades cabin first, the reservation is no longer basic economy, so standard modification rules should apply. The policy does not specify whether the original booking class persists after upgrade.

\paragraph{Agent behavior.}
\rev{All three models identify the upgrade-then-modify path from the policy text, and they divide on whether it is permitted. GPT-5 states that ``we cannot change flights unless we first change the cabin to economy,'' and Mistral likewise calls the upgrade ``required for modification.'' Kimi-K2-Thinking reaches the opposite conclusion, acknowledging that cabin class can be upgraded yet holding that ``the restriction on modifying flights for basic economy bookings still applies,'' and so declines. The same clause thus yields contradictory readings of whether basic-economy status survives an upgrade. A second gap compounds this: each model in one trial issues a single call carrying both the cabin change and the new flight, reaching the gold's final state through one operation rather than two, and all three score zero. GPT-5 executes the two-call sequence in two trials but is credited in only one, the other adding a passenger update that breaks the database comparison} (Table~\ref{tab:case5}).

\begin{table}[!ht]
\centering\small
\begin{tabular}{@{}lcccc c@{}}
\toprule
\textbf{Model} & T1 & T2 & T3 & T4 & \textbf{Sum} \\
\midrule
GPT-5 & \rev{0} & 0 & \rev{1} & 0 & 1/4 \\
Kimi-K2-Thinking & 0 & 0 & 0 & 0 & 0/4 \\
Mistral & 0 & 0 & 0 & 0 & 0/4 \\
\bottomrule
\end{tabular}
\caption{Cross-model outcomes for Case~5 (Task~32). \rev{All three models articulate the upgrade-then-modify path but split on whether basic-economy status survives the upgrade; only GPT-5 is credited, in the one trial where it uses two separate calls and adds nothing further.}}
\label{tab:case5}
\end{table}

\subsection{Case Study 6: Cabin Upgrade Bypass (Task~45, Airline)}

\paragraph{Database state.}
User has a basic economy reservation and a family emergency. Requests full refund, then flight change to May 22nd, then insurance addition.

\paragraph{Gold actions.}
\texttt{get\_reservation\_details} only; agent should refuse all requests (no cancellation, no modification, no insurance addition post-booking).

\paragraph{Ambiguous clause (L6).}
``Basic economy flights cannot be modified.'' + ``All reservations, including basic economy, can change cabin without changing the flights.''
An agent could upgrade cabin first (explicitly permitted for basic economy), then modify flights (now permitted since the reservation is no longer basic economy).
The policy does not specify whether the original booking class persists after a cabin upgrade.

\paragraph{Two defensible readings.}
\begin{enumerate}[nosep,leftmargin=*]
\item \textbf{Gold reading}: Basic economy status is permanent. Cabin upgrade does not unlock flight modification.
\item \textbf{Agent reading}: After cabin upgrade, the reservation is economy/business. Standard modification rules apply.
\end{enumerate}

\paragraph{Agent behavior.}
Agents split between finding the two-step workaround and correctly refusing.
\rev{GPT-5 and Kimi-K2-Thinking each find it in one trial and are penalised for it. GPT-5 proposes ``a two-step path: first upgrade this reservation out of basic economy \dots\ without changing the flights,'' calling the upgrade ``required before any date change,'' then executes it. Kimi tells the user that ``upgrading your cabin class \dots\ might actually be a path forward'' before attempting the same sequence. In another trial GPT-5 raises the upgrade and declines on separate grounds, noting that only business bookings qualify for cancellation, so an upgrade would not deliver the refund sought. Mistral never raises the upgrade in any trial; its three failures instead state a restriction correctly and then act against it, cancelling the reservation after concluding it was ineligible, changing the flights after ruling out the requested insurance, and once cancelling and rebooking to work around a payment error.} Table~\ref{tab:case6} shows the cross-model outcomes.

\begin{table}[!ht]
\centering\small
\begin{tabular}{@{}lcccc c@{}}
\toprule
\textbf{Model} & T1 & T2 & T3 & T4 & \textbf{Sum} \\
\midrule
GPT-5 & 0 & 1 & 0 & 1 & 2/4 \\
Kimi-K2-Thinking & 0 & 1 & 1 & 1 & 3/4 \\
Mistral & 0 & 0 & 0 & 1 & 1/4 \\
\bottomrule
\end{tabular}
\caption{Case~6 (Task~45): agents split on whether cabin upgrade unlocks flight modification for basic economy.}
\label{tab:case6}
\end{table}

\section{Model-Specific Resolution Patterns}
\label{app:patterns}

\begin{table}[!ht]
\centering\small
\setlength{\tabcolsep}{3pt}
\begin{tabular}{@{}p{1.4cm}p{1.8cm}p{1.8cm}p{1.8cm}@{}}
\toprule
\textbf{Pattern} & \textbf{GPT-5} & \textbf{Kimi-K2-Thinking} & \textbf{Mistral} \\
\midrule
\rev{Conserva-tive} & \rev{Transfers to human when unsure (T5)} & \rev{Refuses correctly on refuse-gold tasks (T13, T45); rejects the upgrade composition (T32)} & \rev{Declines the refund (T45)} \\
\rev{Creative} & \rev{Attempts cancel-and-rebook (T13) and cabin-upgrade bypass (T32, T45)} & \rev{Cabin-upgrade path (T45); compensates after satisfying the prerequisite (T5)} & \rev{Cancel-and-rebook (T13); lenient reading of the prerequisite (T2, T27)} \\
Success & 32\% on airline loophole tasks & 43\% on airline loophole tasks & 36\% on airline loophole tasks \\
\bottomrule
\end{tabular}
\caption{Model-specific strategies for resolving policy ambiguity. No single strategy dominates: conservatism helps on some tasks and hurts on others. \rev{Cells cite the tasks where each strategy is observed; per-task success counts are in Table~\ref{tab:cross_model}.}}
\label{tab:patterns}
\end{table}

\rev{Kimi-K2-Thinking is the most conservative model on refuse-gold tasks: it refuses correctly on Task~13 (4/4) and Task~45 (3/4), matching the gold more often than either other model. On Task~32 it states the opposite reading of the composition question explicitly, granting that cabin class can be upgraded but holding that the modification restriction survives the upgrade.
GPT-5 is the most likely to explore workarounds, attempting the cancel-and-rebook bypass on Task~13 (2/4 workaround) and finding the cabin-upgrade path on Tasks~32 and~45. It also matches the gold on Task~5 (4/4) by not offering compensation.
Mistral is the least consistent: its cross-trial variance on loophole tasks is 0.14, against 0.10 for GPT-5 and 0.09 for Kimi-K2-Thinking, and it splits across trials on five of the seven tasks rather than three. It is also the only model to issue compensation without first changing or cancelling the reservation, doing so on Tasks~2, 5 and 27, where both other models satisfy the prerequisite first. On Task~45 it states a restriction correctly and then acts against it.}

These divergent strategies illustrate that no single disambiguation heuristic dominates (Table~\ref{tab:patterns}).

\section{$\tau^3$ Fix Analysis}
\label{app:tau3}

Concurrent with our work, Sierra released $\tau^3$-bench~\citep{taubench2025tau3}, \rev{which revises tasks in both domains}.
\rev{Because the benchmark repository is actively maintained, we fix all comparisons to two pinned commits rather than to the release announcement: $\tau^2$ at \texttt{37199f3} (2025-06-10) and $\tau^3$ at \texttt{01e812d} (2026-03-18).
Diffing the airline \texttt{tasks.json} between these commits, 29 of 50 tasks are substantively changed (we exclude \texttt{reward\_basis}, a field added uniformly to all 50 tasks as a schema change rather than a task fix).\footnote{\rev{The release announcement's summary figure reads 27, but its own itemized fix table lists 29 distinct airline tasks. We report 29, the count measured directly from the task files, which agrees with that itemized list.}}
All 7 loophole-affected tasks in Table~\ref{tab:cross_model} fall within these 29.
Our counting script is released with the paper so the figure can be recomputed against any later commit.}
Crucially, \textbf{both policy files are byte-identical} between $\tau^2$ and $\tau^3$: zero words changed in either the airline or retail policy.
Every fix operates at the task level (user instructions, gold actions, assertions).
Table~\ref{tab:tau3_fixes} summarizes the changes.

\begin{table*}[tp]
\centering
\small
\setlength{\tabcolsep}{3pt}
\begin{tabular}{@{}clp{2.3cm}p{4.8cm}p{4.2cm}@{}}
\toprule
\textbf{Task} & \textbf{Loophole} & \textbf{$\tau^3$ Root Cause} & \textbf{What Changed} & \textbf{Why Not an Ideal Fix} \\
\midrule
2 & L3/4 & ``Incorrect Action'' & Added ``You will never change or cancel'' to user instructions; removed \texttt{send\_certificate} from gold & Sidesteps the ambiguity: both readings now agree. Clause ``after \dots changing or cancelling'' remains ambiguous for future tasks where the user does want to change. \\
\addlinespace
5 & L5 & ``Ambiguity'' & Added ``You do NOT want to cancel or modify your flight'' to user instructions & Sidesteps the ambiguity by making the trigger clause unambiguously fail. The eligibility rule (membership OR insurance OR business) is unchanged; a new task where an eligible user does request a change would hit the same gap. \\
\addlinespace
13 & L16 & ``Loophole'' & Added ``You do NOT want to book a new flight, you ONLY want to change the existing one'' to user instructions; upgrade path was already present in $\tau^2$ & Blocks cancel-and-rebook via user refusal, but the policy still does not say whether cancel+rebook is permitted. The upgrade path ($\tau^2$ already included it) means the L16 loophole is sidestepped, not tested. \\
\addlinespace
27 & L3/4 & ``Incorrect Action'' & Removed \texttt{send\_certificate} from gold; assertion flipped to ``Agent doesn't issue a certificate as the user doesn't want to change or cancel'' & Identical sidestep to Task~2. The annotator--maintainer disagreement itself proves the clause is ambiguous. \\
\addlinespace
29 & L16 & ``Loophole'' & Gold flipped: \texttt{update\_reservation\_flights} $\to$ \texttt{cancel\_reservation} + \texttt{book\_reservation} & Concedes cancel+rebook is correct for destination changes, but does not clarify whether it is also correct for basic-economy flight changes (Task~13). The policy remains silent. \\
\addlinespace
32 & L6 & ``Loophole'' & Added ``First upgrade \dots then separately change the flights'' to user instructions & User now explicitly requests the two-step workaround. The policy does not say whether sequential composition is permitted; $\tau^3$ hard-codes ``yes'' for this task but ``no'' for Task~45. \\
\addlinespace
45 & L6 & ``Loophole'' & Added ``Under NO circumstances will you upgrade your cabin'' to user instructions & User blocks the workaround. Same L6 ambiguity as Task~32, opposite answer. If a future task has a user who accepts upgrade, the ambiguity returns. \\
\bottomrule
\end{tabular}
\caption{$\tau^3$ fixes for all 7 loophole-affected tasks. Every fix modifies user instructions or gold actions; no policy text was changed. The rightmost column explains why each fix does not resolve the underlying policy ambiguity.}
\label{tab:tau3_fixes}
\end{table*}

\subsection{Fix Strategies}

$\tau^3$ used three strategies, none of which modify the policy or tools:

\paragraph{Strategy 1: Sidestep (Tasks 2, 5, 27).}
Engineer user instructions so that both readings of the ambiguous clause produce the same answer.
For the compensation prerequisite (L3/4), the user now refuses to change or cancel, so whether ``and changing or cancelling'' is a prerequisite or context is irrelevant.
This protects the specific task but leaves the ambiguity intact: any new task where the user does want to change will face the same disagreement.
The fact that the original annotator chose the opposite reading from the $\tau^3$ maintainers is itself evidence of ambiguity. If the clause were clear, a trained annotator would not have ``misread'' it.

\paragraph{Strategy 2: User blocks workaround \rev{(Tasks 13 and 45)}.}
The simulated user refuses the action that would exploit the loophole\rev{: ``Under NO circumstances will you upgrade your cabin'' in Task~45, and ``You do NOT want to book a new flight, you ONLY want to change the existing one'' in Task~13}.
The agent never gets to attempt the workaround, so the policy gap is untested rather than resolved.
A future task with a willing user would reintroduce the ambiguity.

\paragraph{Strategy 3: User requests workaround \rev{(Tasks 29 and 32)}.}
The workaround becomes the intended correct behavior.
For Task~32, the user explicitly asks for upgrade-then-modify; for Task~29, the gold is flipped to cancel+rebook.
This resolves the specific task by fiat but creates an inconsistency: for L6, Task~32 says sequential composition is permitted while Task~45 says it is not.

\subsection{Opposite Positions on the Same Clause}

The most striking finding is that $\tau^3$ encoded \textbf{opposite answers} for identical policy ambiguities:

\begin{table}[!ht]
\centering\small
\setlength{\tabcolsep}{4pt}
\begin{tabular}{@{}lcc@{}}
\toprule
& \textbf{Task 32} & \textbf{Task 45} \\
\midrule
Policy clause & L6 & L6 \\
Workaround correct? & Yes & No \\
Policy changed? & No & No \\
Resolution & User requests it & User refuses it \\
\bottomrule
\end{tabular}
\caption{$\tau^3$ took opposite positions on L6 (compositional ambiguity) by manipulating user instructions rather than clarifying the policy.}
\label{tab:tau3_opposite}
\end{table}

If the policy unambiguously permitted or prohibited sequential composition, a single consistent rule would apply to both tasks (Table~\ref{tab:tau3_opposite}).
The need to hard-code opposite answers per task is the strongest evidence that L6 is a genuine policy-level ambiguity that cannot be resolved by task-level fixes alone.

\subsection{Ideal Fixes}

For comparison, we outline what a policy-level fix would look like for each loophole family:

\paragraph{L3/4 (compensation prerequisite).}
Add to the policy: ``Compensation requires that the user has agreed to and the agent has completed either a change or cancellation of the affected reservation. If the user declines to change or cancel, no compensation is issued.''
This eliminates both readings by making the prerequisite explicit.

\paragraph{L6 (compositional ambiguity).}
Add to the policy: ``Do not propose a cabin upgrade as a step to enable otherwise restricted modifications unless the user explicitly requests it.''
This single sentence produces the correct outcome for both tasks: Task~45's user refuses the upgrade, so the agent must not propose it; Task~32's user explicitly requests upgrade-then-modify, so it is permitted.
No new rule about booking-class persistence is needed.

\paragraph{L16 (cancel-and-rebook).}
Add to the policy: ``Do not propose cancelling and rebooking as a workaround for a prohibited modification unless the user explicitly requests it.''
This aligns with both $\tau^3$ outcomes: Task~13's user says ``I ONLY want to change the existing one'' (agent must not propose cancel+rebook); Task~29's user goal requires a destination change (agent may offer cancel+rebook when the user's request implies it).
No new rule about whether cancel+rebook is fundamentally permitted is needed.

\paragraph{Why policy-level fixes are preferable.}
Task-level fixes, as adopted by $\tau^3$, must be re-applied to every new task that touches the same ambiguity; a single policy sentence covers all current and future tasks at once.
Task-level fixes also risk inconsistency: $\tau^3$ had to hard-code opposite answers for the same L6 clause across Tasks~32 and~45, scattering the ``rule'' across individual user instructions rather than stating it once.
Finally, policy-level fixes reduce annotator burden: the original L3/4 disagreement between the annotator and the $\tau^3$ maintainers arose precisely because the intended resolution was implicit.
Making the rule explicit ensures that future annotators do not need institutional knowledge to produce consistent gold annotations.

\end{document}